%% file: header.tex
\documentclass{article}

\usepackage{microtype}
\usepackage{graphicx}
\usepackage{booktabs} 
\usepackage{siunitx}   
\usepackage{tabularx}
\usepackage{subcaption}

\usepackage{hyperref}

\usepackage[accepted]{icml2025}

\usepackage{amsmath}
\usepackage{amssymb}
\usepackage{mathtools}
\usepackage{amsthm}
\usepackage{mdframed}
\usepackage{xcolor}

\usepackage[capitalize,noabbrev]{cleveref}

\theoremstyle{plain}

\theoremstyle{definition}

\theoremstyle{remark}

\usepackage[textsize=tiny]{todonotes}

\usepackage[export]{adjustbox}
\usepackage{xspace}
\usepackage[normalem]{ulem}
\useunder{\uline}{\ul}{}
\usepackage[most]{tcolorbox}
\usepackage{wrapfig}

\def\ie{\emph{i.e}.,\xspace}

\icmltitlerunning{On the Effect of Uncertainty on Layer-wise Inference Dynamics}

\begin{document}

\twocolumn[
\icmltitle{On the Effect of Uncertainty on Layer-wise Inference Dynamics}



\icmlsetsymbol{equal}{*}

\begin{icmlauthorlist}
\icmlauthor{Sunwoo Kim}{kaist}
\icmlauthor{Haneul Yoo}{kaist}
\icmlauthor{Alice Oh}{kaist}
\end{icmlauthorlist}

\icmlaffiliation{kaist}{School of Computing, KAIST, Daejeon, South Korea}

\icmlcorrespondingauthor{Sunwoo Kim}{sunwoo.kim@kaist.ac.kr}
\icmlcorrespondingauthor{Haneul Yoo}{haneul.yoo@kaist.ac.kr}
\icmlcorrespondingauthor{Alice Oh}{alice.oh@kaist.edu}

\icmlkeywords{Mechanistic Interpretability, Uncertainty Quantification, Large Language Models}

\vskip 0.3in
]



\printAffiliationsAndNotice{}  

\begin{abstract}
\input{texes/0_abstract}
\end{abstract}

\section{Introduction}
\input{texes/1_introduction}

\section{Using Tuned Lens to Extract Inference Dynamics}
\input{texes/3_method}

\section{Experimental Results}
\input{texes/4_result}

\section{Related Work}
\input{texes/2_related_works}

\section{Conclusion}
\input{texes/5_conclusion}

\section*{Acknowledgements}
This work was supported by Institute of Information \& communications Technology Planning \& Evaluation(IITP) grant funded by the Korea government(MSIT) (No. RS-2024-00509258 and No. RS-2024-00469482, Global AI Frontier Lab).
This work was supported by Institute of Information \& communications Technology Planning \& Evaluation (IITP) grant funded by the Korea government(MSIT) (No.RS-2022-II220184, Development and Study of AI Technologies to Inexpensively Conform to Evolving Policy on Ethics).



\bibliography{bibliography}
\bibliographystyle{icml2025}

\newpage
\appendix
\onecolumn
\section*{Appendix}
\input{texes/appendix}

\end{document}

%% file: texes/0_abstract.tex
Understanding how large language models (LLMs) internally represent and process their predictions is central to detecting uncertainty and preventing hallucinations. 
While several studies have shown that models encode uncertainty in their hidden states, it is underexplored how this affects the way they process such hidden states. 
In this work, we demonstrate that the dynamics of output token probabilities across layers for certain and uncertain outputs are largely aligned, revealing that uncertainty does not seem to affect inference dynamics. 
Specifically, we use the Tuned Lens, a variant of the Logit Lens, to analyze the layer-wise probability trajectories of final prediction tokens across 11 datasets and 5 models. 
Using incorrect predictions as those with higher epistemic uncertainty, our results show aligned trajectories for certain and uncertain predictions that both observe abrupt increases in confidence at similar layers. 
We balance this finding by showing evidence that more competent models may learn to process uncertainty differently.
Our findings challenge the feasibility of leveraging simplistic methods for detecting uncertainty at inference. 
More broadly, our work demonstrates how interpretability methods may be used to investigate the way uncertainty affects inference. 

%% file: texes/1_introduction.tex
As capacities of LLMs grow, so does users' reliance on them. Hence, it is important to understand models' awareness of uncertainty and be able to detect it. To this end, many researchers have leveraged interpretability methods to study uncertainty calibration and quantification.
Research using black-box methods have shown that models reflect uncertainty in their outputs~\cite{kadavath2022language,wang2024ubenchbenchmarkinguncertaintylarge}. Mechanistic interpretability (MI) methods, such as using sparse auto-encoders~\cite{ferrando2025do} and linear probes on embeddings of frozen, pretrained models~\cite{ahdritz2024distinguishing}, have also found that models encode features for uncertainty in their internal representations.
Although existing MI analyses provide ample evidence that \textit{what} the model processes (\ie embeddings) is different for uncertain outputs, the way this affects \textit{how} models process uncertain outputs differently is underexplored. 
Models could adapt their inference dynamics to uncertainty, perhaps processing outputs with greater uncertainty more, in terms of layers effectively used. 
If so, we may leverage these distinctly different inference dynamics for uncertainty detection. 
Additionally, it would seem natural that more competent models would exhibit more adaptive ways of processing uncertain predictions. 
Motivated by the thoughts above, we present two exploratory research questions:
\begin{itemize}
    \item \textbf{RQ1: }
    How does uncertainty affect the inference dynamics of models? (\emph{Figure~\ref{fig:trajectories}, Figure~\ref{fig:pd_dist}})
    \item \textbf{RQ2: }
    Does model competence affect the ability to adapt its inference dynamics to uncertainty? (\emph{Figure~\ref{fig:kappa}})
\end{itemize}

To answer the research questions, we analyze inference dynamics by using the levels of confidence the model has on possible prediction tokens throughout its layers. 
We extract such confidence levels from the hidden states passed in the residual stream between layers by employing the Tuned Lens~\cite{belrose2023eliciting}, a variant of the Logit Lens~\cite{nostalgebraist2020logit}. We gather data across 11 datasets with 5 LLMs. We focus on \emph{epistemic uncertainty}~\cite{ahdritz2024distinguishing, hou2024decomposing}, uncertainty due to lack of a model's knowledge and training, which is more easily observable compared to its counterpart, \emph{aleatoric uncertainty}; the model exhibits epistemic uncertainty when it incorrectly answers a question from a set of answer choices that includes the correct answer. 

Through our extensive experiments, we observe that probability trajectories are strikingly aligned and that models decide on their outputs at certain layers largely independent of uncertainty. 
This inference pattern means more complex and nuanced interpretability approaches may be required to extract model uncertainty at inference. 
While \citet{jiang-etal-2024-large} examined inference dynamics in the context of hallucination of known facts, our research extends this understanding across diverse tasks and models for epistemic uncertainty. 
Further, we add nuance to our findings by showing evidence that adaptive inference dynamics may arise with greater model competence.

%% file: texes/3_method.tex
\input{src/figure/trajectories}
\input{src/figure/pd_dist}

We use the Tuned Lens~\cite{belrose2023eliciting} to convert the hidden states in the residual stream after each layer to logit distributions in the vocabulary space. The logit distributions are indicative of what the model \emph{believes} is the correct prediction after a layer. 
We perform this analysis on single token answers for multiple-choice question (MCQ) datasets. 
We analyze the distribution data to obtain tokens' probability trajectory through the layers, as well as the prediction depth---the layer at which the model commits to its answer. 

\subsection{Probability Trajectory Analysis}
We analyze how the probabilities of possible predictions, namely single letter answer labels for the MCQs, change across layers. To clarify, we extract the probability distribution in the possible prediction space by performing softmax on just the logits for the label tokens. Then, we plot the probabilities for each label across layers to obtain the probability trajectories. We aggregate these probability trajectories across questions for all the datasets. Notably, we aggregate the trajectories for questions correctly and incorrectly answered separately.  Specifically, for each question, we extract the trajectory for the label ranked first in terms of its probability at the final layer and obtain its average trajectory across questions. We condense the remaining lower ranked labels into a single trajectory. Hence, we obtain four average trajectories for each model, one pair of top  and low  rank trajectories for correctly answered questions and another for incorrectly answered questions. We compare the trajectories for correctly and incorrectly answered questions to see how inference dynamics differ. 

\subsection{Prediction Depth Analysis}
The layer at which the model commits to its answer, or the prediction depth as coined by \citet{baldock2021deep}, would be the one at which a model's top prediction is different from the previous layer's and is maintained for subsequent layers. 
We calculate the PD for all questions in all datasets. 
We then plot the aggregate percentage distribution of prediction depths for all datasets across layers to see if distributions for when the model commits to its outputs diverge.
Additionally, we calculate the Pearson correlation of answer incorrectness with PD for model and dataset pairs to see if uncertainty affects PD.
Further, we analyze how the Kappa value---accuracy adjusted for chance---on a dataset relates to the average PD difference between correct and incorrect outputs on that dataset. A greater PD difference would mean a greater degree of adaptive layer usage. Through this analysis, we can determine if sensitivity of inference dynamics to uncertainty changes according to task proficiency and, by implication, model competence. A positive correlation would mean that such abilities may emerge to greater degrees as a model's capability grows. 

\subsubsection{Models and Datasets}
We use three models with Tuned Lenses provided by the authors of the Tuned Lens paper~\cite{belrose2023eliciting}: \texttt{Llama-3-8B}, \texttt{Llama-3-8B-Instruct}~\cite{grattafiori2024llama3}, and \texttt{vicuna-13b-v1.1}~\cite{chiang2023vicuna}. Additionally, we train new Tuned Lenses on two models, \texttt{Mistral-7B-Instruct-v0.1}~\cite{jiang2023mistral7b} and \texttt{Mistral-Nemo-Instruct-2407}~\cite{mistral2024nemo}.

We use datasets covering a range of tasks: ANLI~\cite{nie-etal-2020-adversarial}, ARC~\cite{clark2018think}, BoolQ~\cite{clark-etal-2019-boolq}, CommonsenseQA~\cite{talmor-etal-2019-commonsenseqa}, HellaSwag~\cite{zellers-etal-2019-hellaswag}, LogiQA~\cite{liu2021logiqa}, MMLU~\cite{hendrycks2021measuring}, QASC~\cite{khot2020qasc}, QuAIL~\cite{rogers2020getting}, RACE~\cite{lai-etal-2017-race}, and SciQ~\cite{welbl-etal-2017-crowdsourcing}. 

%% file: src/figure/trajectories.tex
\begin{figure*}
    \centering
    \begin{subfigure}{0.19\linewidth}
        \centering
        \includegraphics[width=\linewidth]{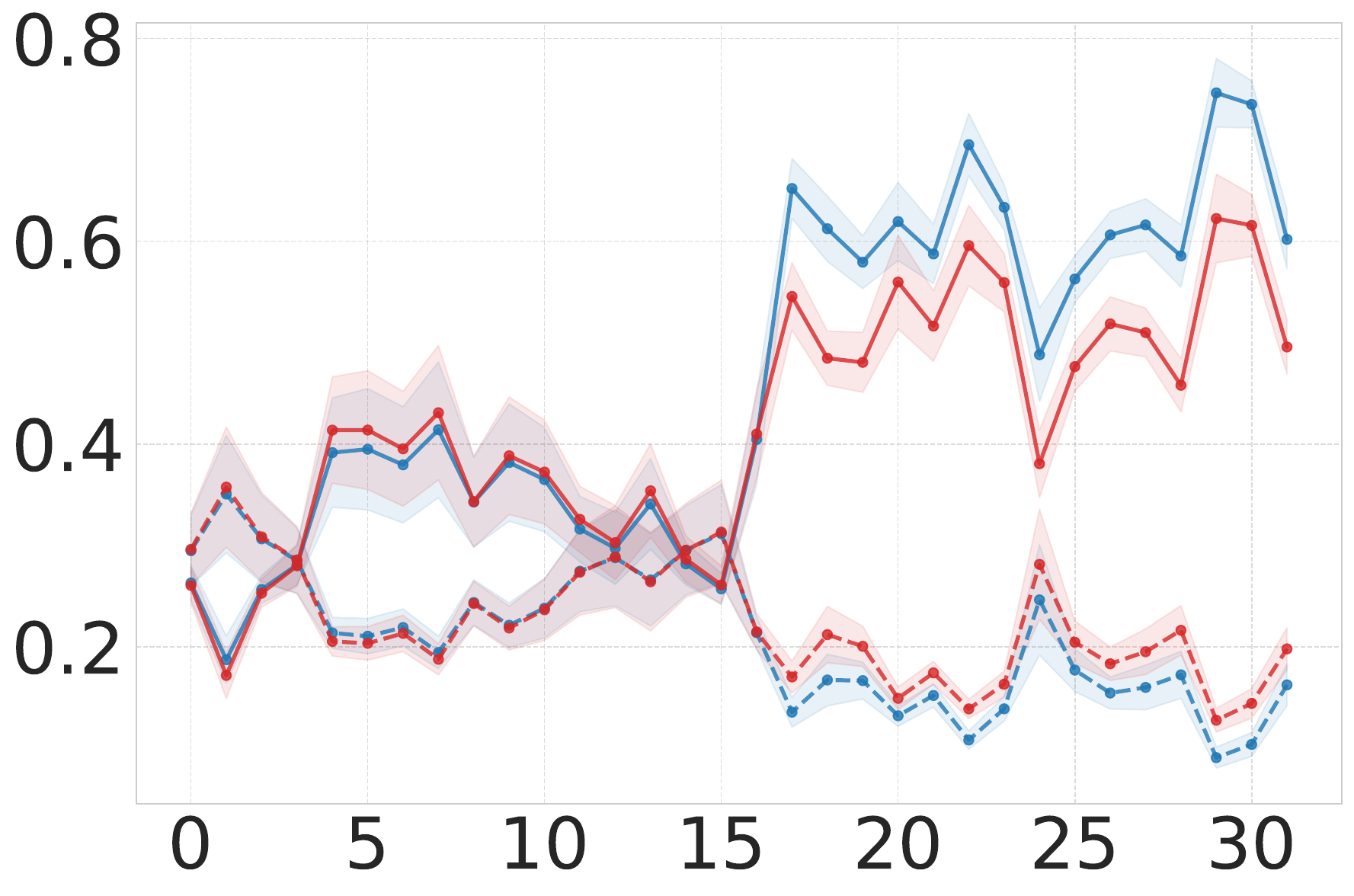}
        \caption{Llama-3-8B}\label{fig:trajectories_llama}
    \end{subfigure}
    \begin{subfigure}{0.19\linewidth}
        \centering
        \includegraphics[width=\linewidth]{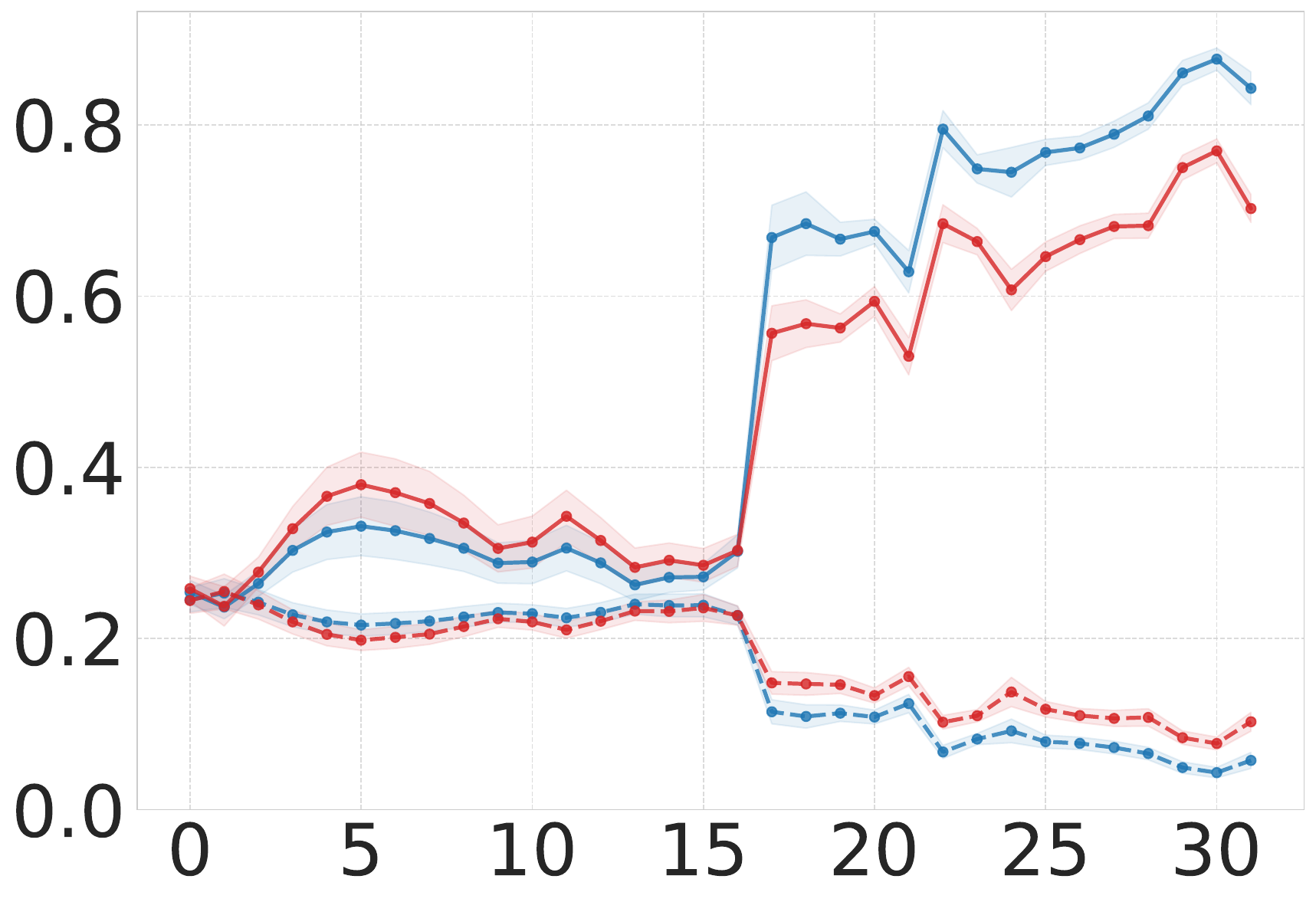}
        \caption{Llama-3-8B-Instruct}\label{fig:trajectories_inst}
    \end{subfigure}
    \begin{subfigure}{0.19\linewidth}
        \centering
        \includegraphics[width=\linewidth]{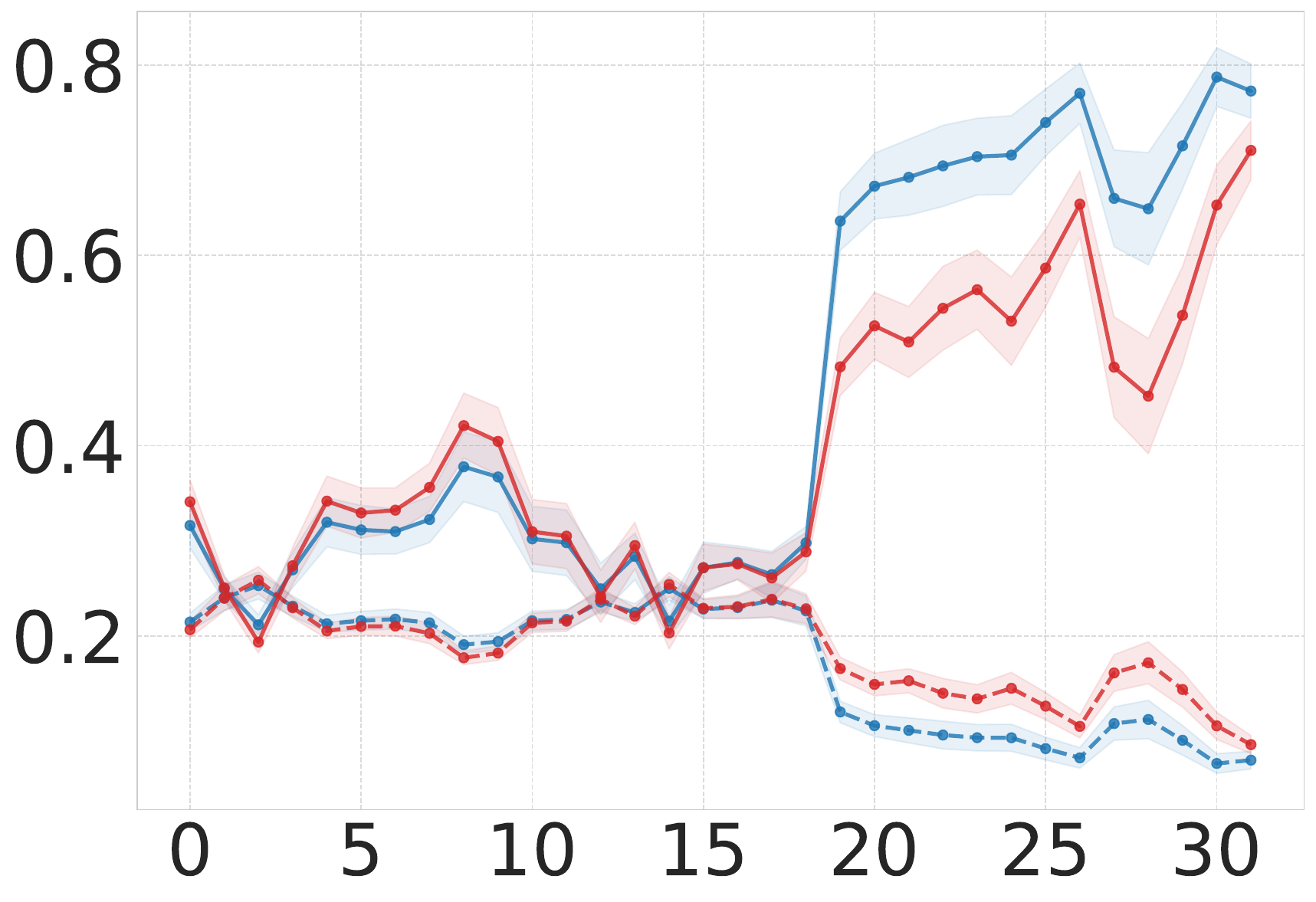}
        \caption{Mistral-7B-Instruct}\label{fig:trajectories_mistral}
    \end{subfigure}
    \begin{subfigure}{0.19\linewidth}
        \centering
        \includegraphics[width=\linewidth]{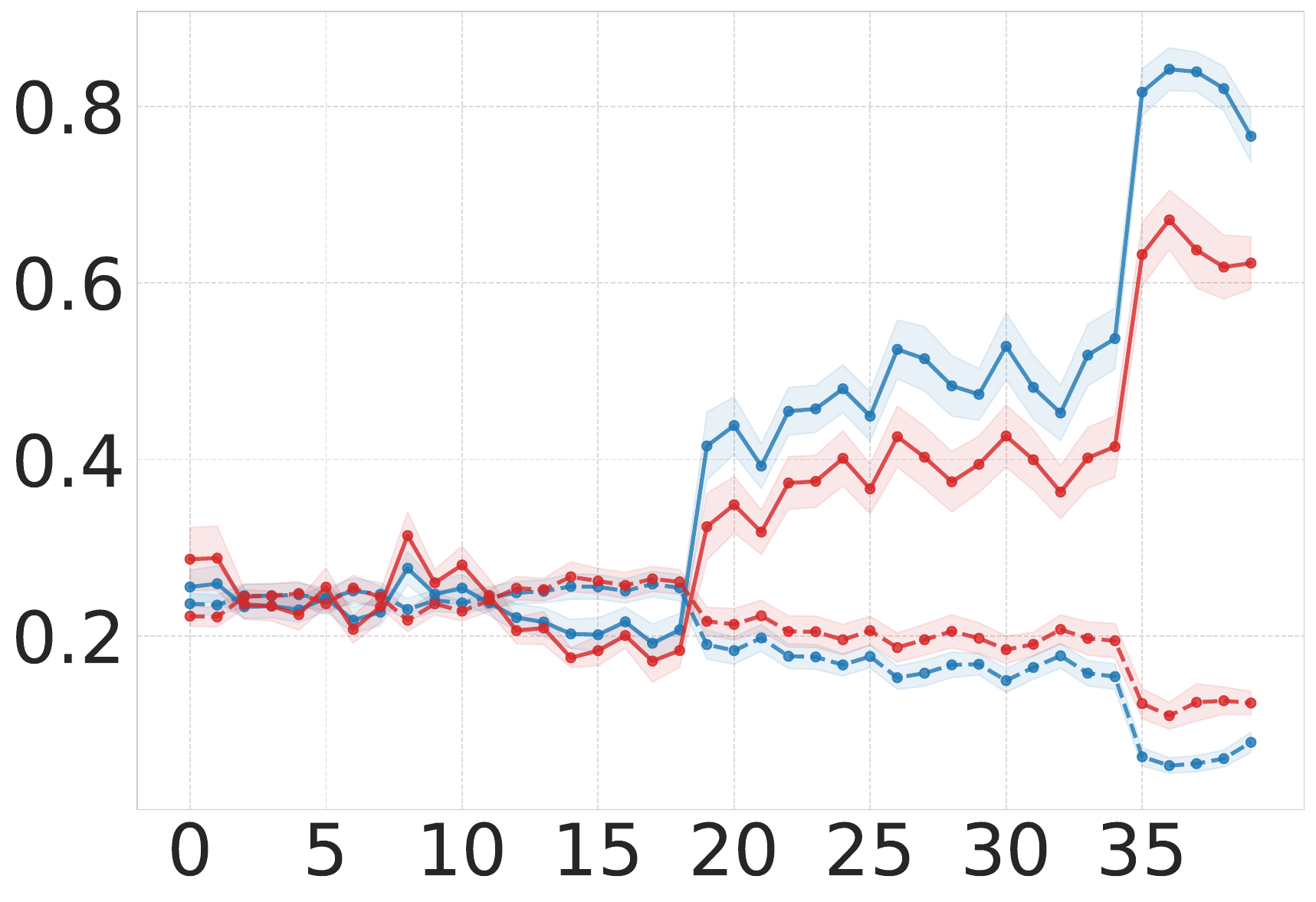}
        \caption{Mistral-Nemo-Instruct}\label{fig:trajectories_mistral_nemo}
    \end{subfigure}
    \begin{subfigure}{0.19\linewidth}
        \centering
        \includegraphics[width=\linewidth]{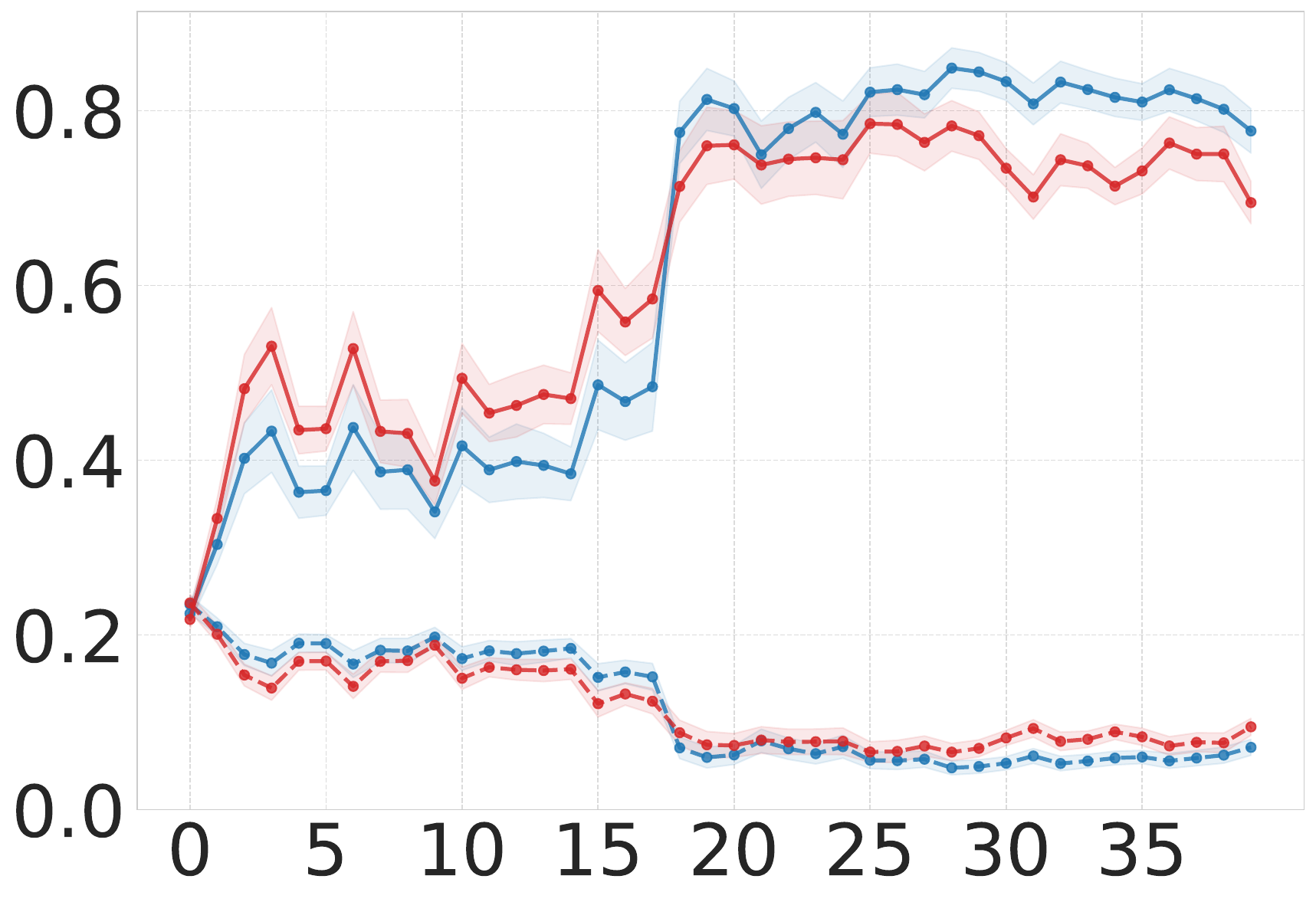}
        \caption{Vicuna-13B}\label{fig:trajectories_vicuna}
    \end{subfigure}
    \caption{Average probability trajectories. The x-axis denotes layer number and the y-axis denotes probability. The plots show how the final prediction token probability (\emph{solid lines}) changes across layers for correct predictions (\emph{in blue}) and incorrect predictions (\emph{in red}). The probabilities for the other possible predictions in each case are condensed in the dashed line. The trajectories show strong alignment, moving synchronously across layers. The model seems to decide on its final prediction, as noted by the abrupt increase in probability, at similar layers as well.}
    \label{fig:trajectories}
\end{figure*}

%% file: src/figure/pd_dist.tex
\begin{figure*}[htb!]
    \centering
    \begin{subfigure}{0.19\linewidth}
        \centering
        \includegraphics[width=\linewidth]{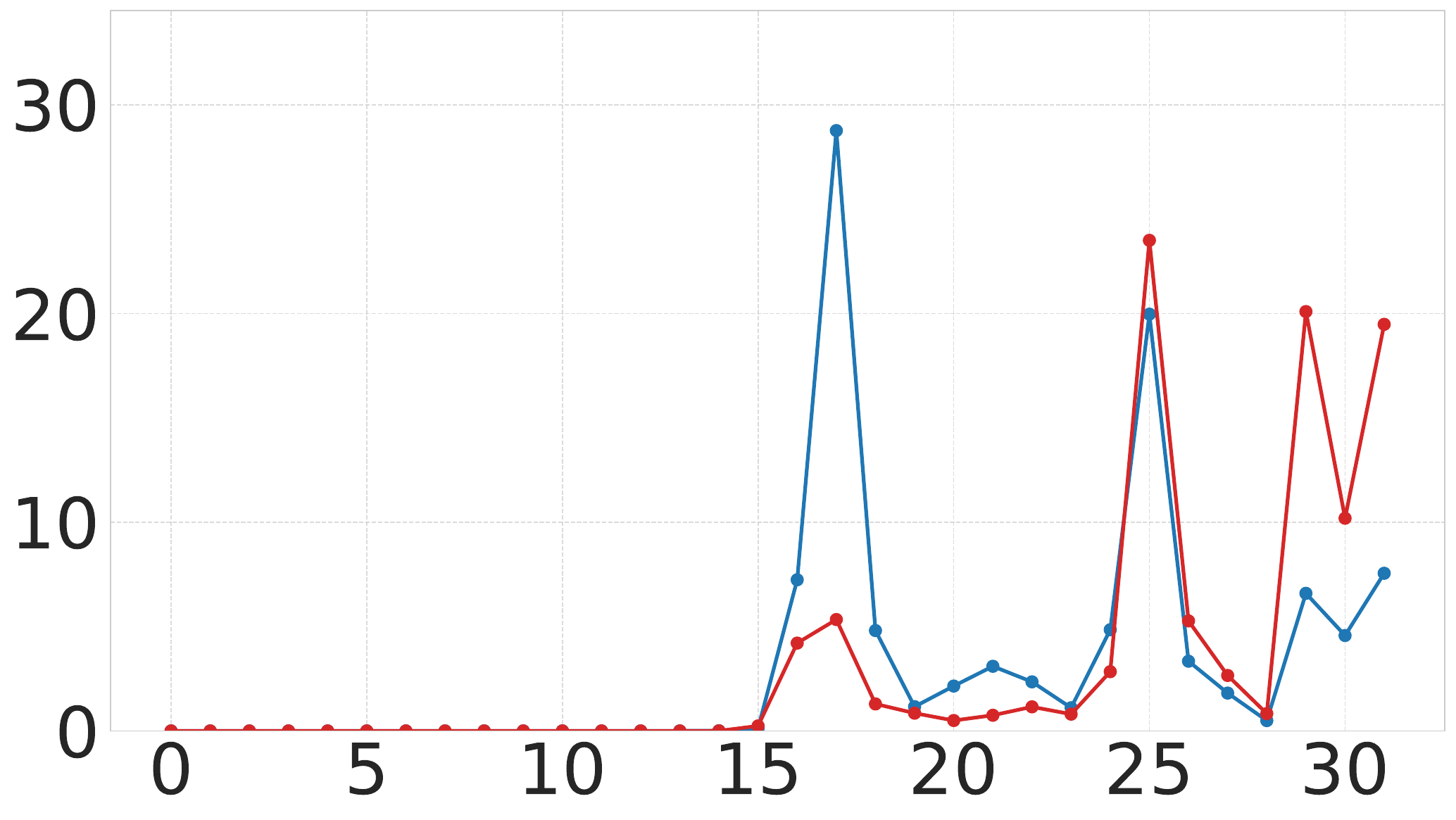}
        \caption{Llama-3-8B}\label{fig:pd_dist_llama}
    \end{subfigure}
    \begin{subfigure}{0.19\linewidth}
        \centering
        \includegraphics[width=\linewidth]{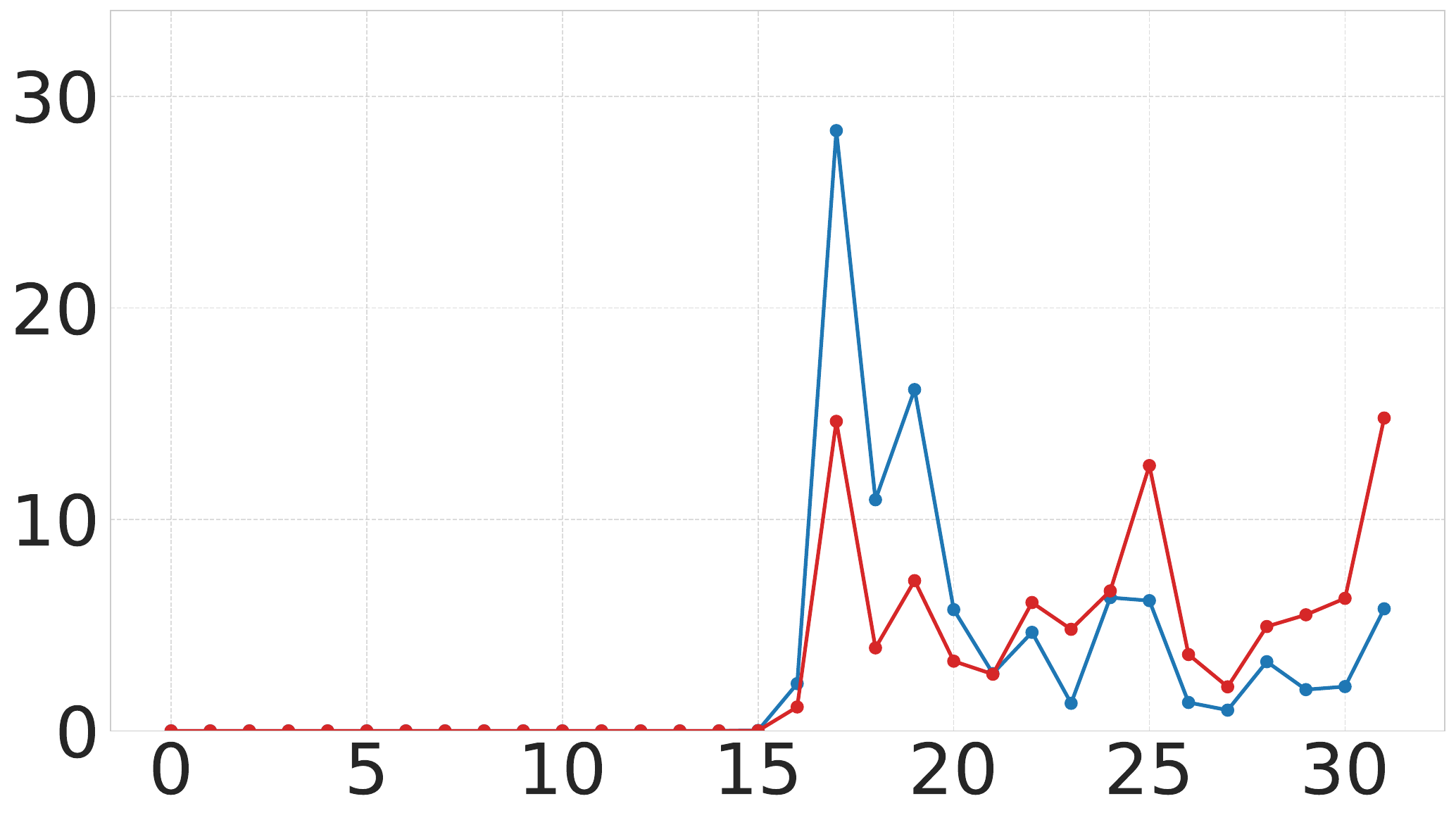}
        \caption{Llama-3-8B-Instruct}\label{fig:pd_dist_inst}
    \end{subfigure}
    \begin{subfigure}{0.19\linewidth}
        \centering
        \includegraphics[width=\linewidth]{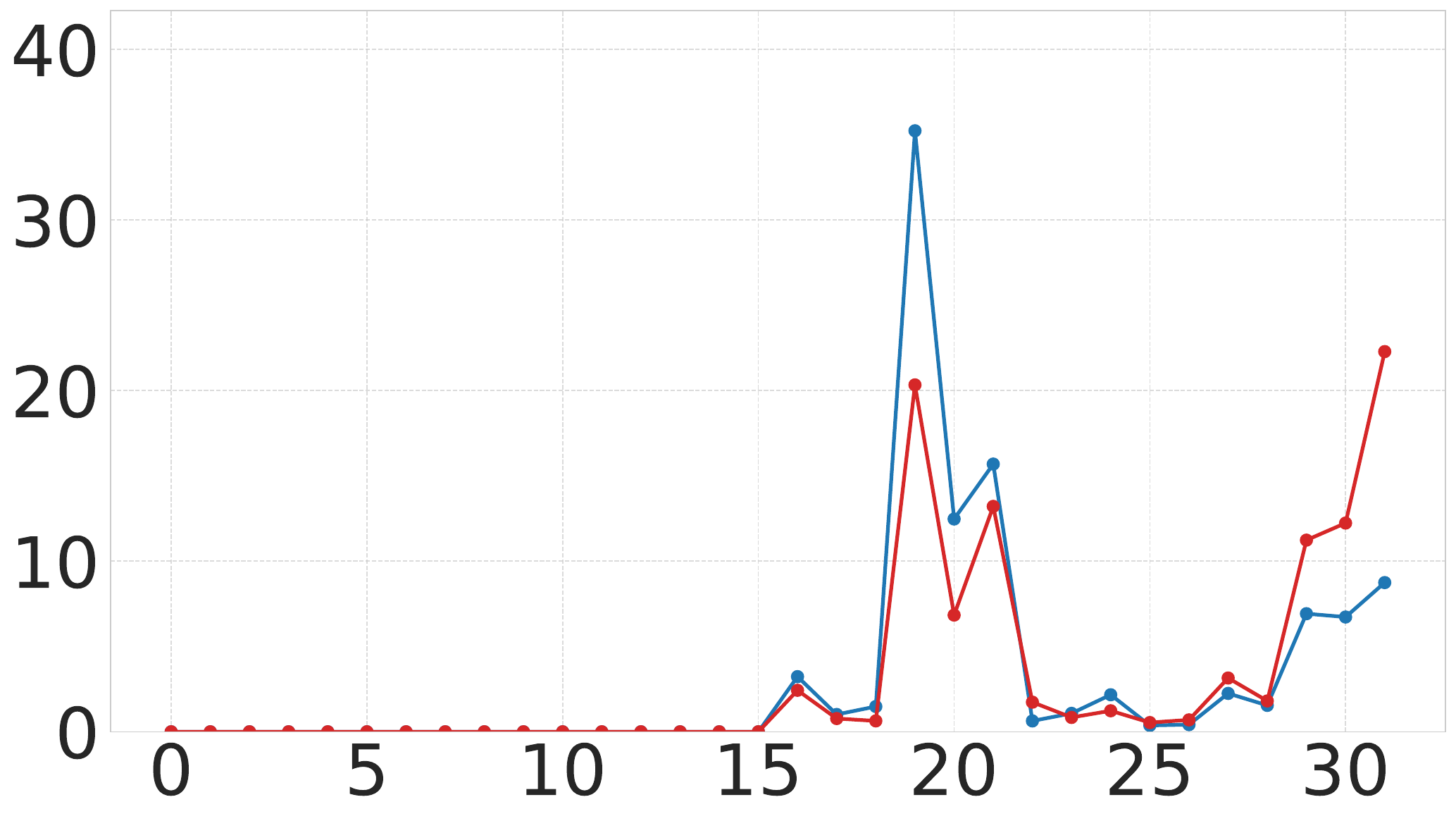}
        \caption{Mistral-7B-Instruct}\label{fig:pd_dist_mistral}
    \end{subfigure}
    \begin{subfigure}{0.19\linewidth}
        \centering
        \includegraphics[width=\linewidth]{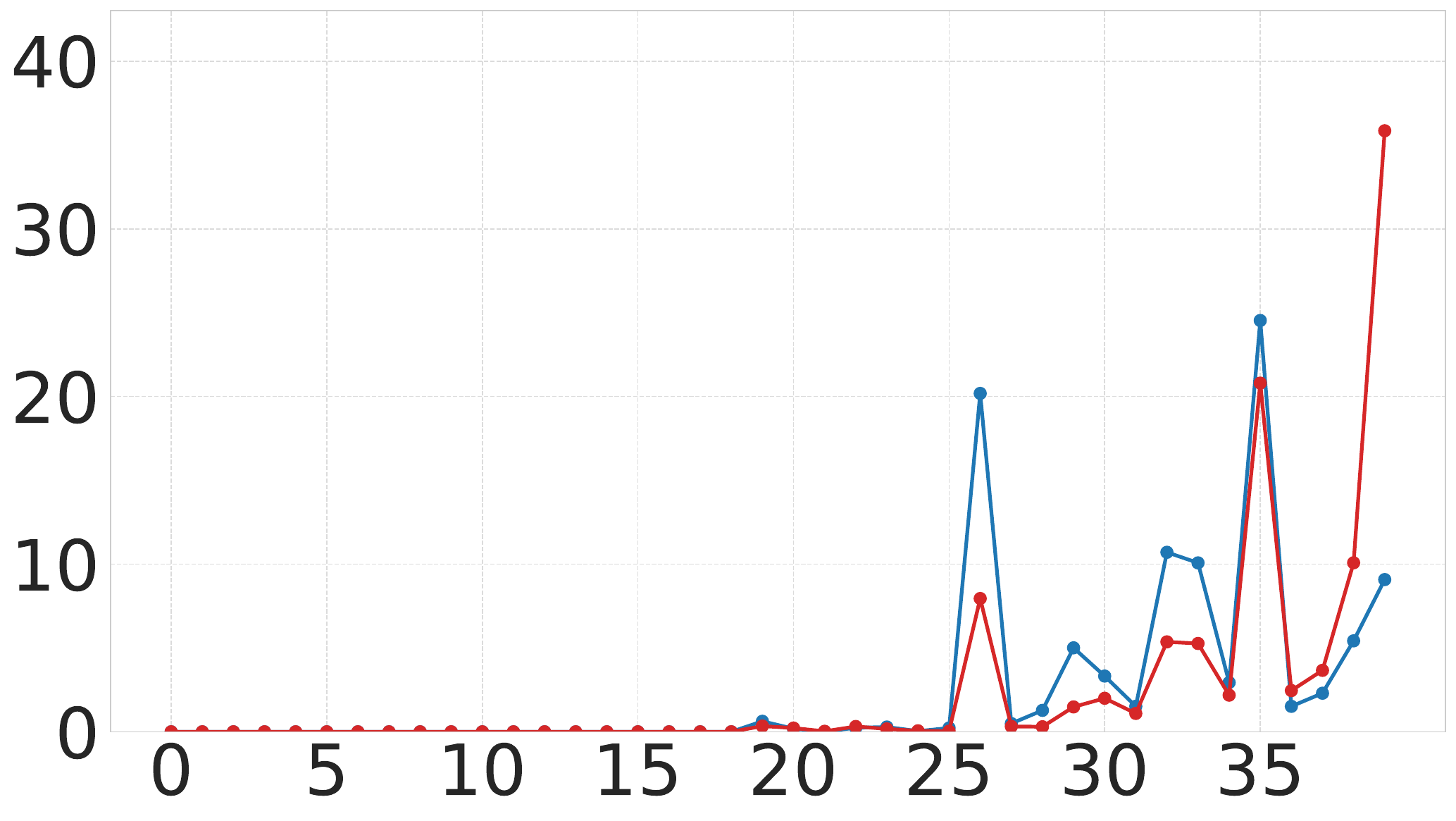}
        \caption{Mistral-Nemo-Instruct}\label{fig:pd_dist_mistral_nemo}
    \end{subfigure}
    \begin{subfigure}{0.19\linewidth}
        \centering
        \includegraphics[width=\linewidth]{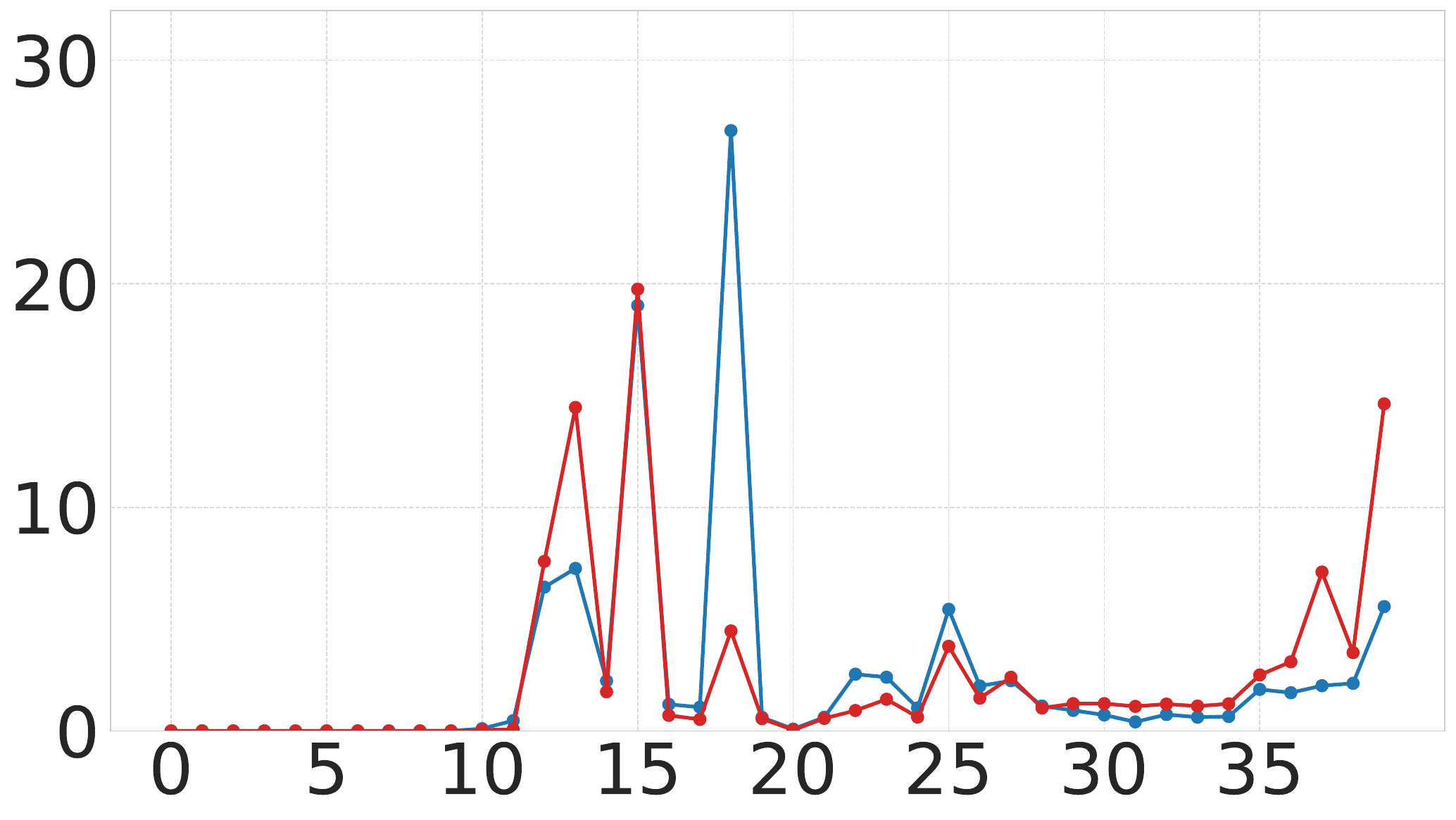}
        \caption{Vicuna-13B}\label{fig:pd_dist_vicuna}
    \end{subfigure}
    \caption{Prediction depth distribution. The x-axis denotes layer number and the y-axis denotes percentage of questions. The plots show how the prediction depth, or the layer at which the model committed to its output, is distributed similarly for correct (\emph{in blue}) and incorrect answers (\emph{in red}). This alignment shows that the model reserves a similar number of layers for processing certain and uncertain outputs.}
    \label{fig:pd_dist}
\end{figure*}

%% file: texes/4_result.tex
Figure~\ref{fig:trajectories} shows a consistent pattern across models in which the top prediction trajectory sharply increases and maintains its divergence with the lower rank trajectory. This is when the model seems to decide on its output.
The trajectories for correct and incorrect predictions are strongly aligned, moving synchronously and abruptly increasing at the same layers. Albeit, the top prediction trajectory for the incorrect questions is consistently lower after the sharp increase, we cannot observe a definitive divergence between the two that would indicate different inference dynamics. 
We can also observe the alignment in the distribution of PD layers in Figure~\ref{fig:pd_dist} where there are similar peaks, meaning the model committed to its final predictions at the same layers regardless of uncertainty.

\input{src/figure/kappa}
\input{src/table/pd_correlation}

PD correlations with answer incorrectness in Table~\ref{tab:pd-corr-table} show weak positive correlations across models and datasets, with 80\% below 0.300. However, the results are 97\% positive. 
This consistency shows the potential for models to leverage adaptive inference techniques.
To investigate this potential specifically in terms of emergence in competent models, we plot how the PD difference relates to the Kappa score for each dataset in Figure~\ref{fig:kappa}. The figure shows positive linear correlations between Kappa scores and the average PD gap for datasets. Two out of the five models, \texttt{Llama-3-8B} and \texttt{Mistral-7B-Instruct}, show statistically significant positive correlations with $p < 0.05$ while \texttt{Llama-3-8B-Instruct} has $p = 0.06$. Insofar as we can interpret accuracy on a dataset as task proficiency, these results are preliminary evidence that stronger adaptive inference dynamics could emerge with greater competence.

%% file: src/figure/kappa.tex
\begin{figure*}[htb!]
    \centering
    \begin{subfigure}{0.19\linewidth}
        \centering
        \includegraphics[width=\linewidth]{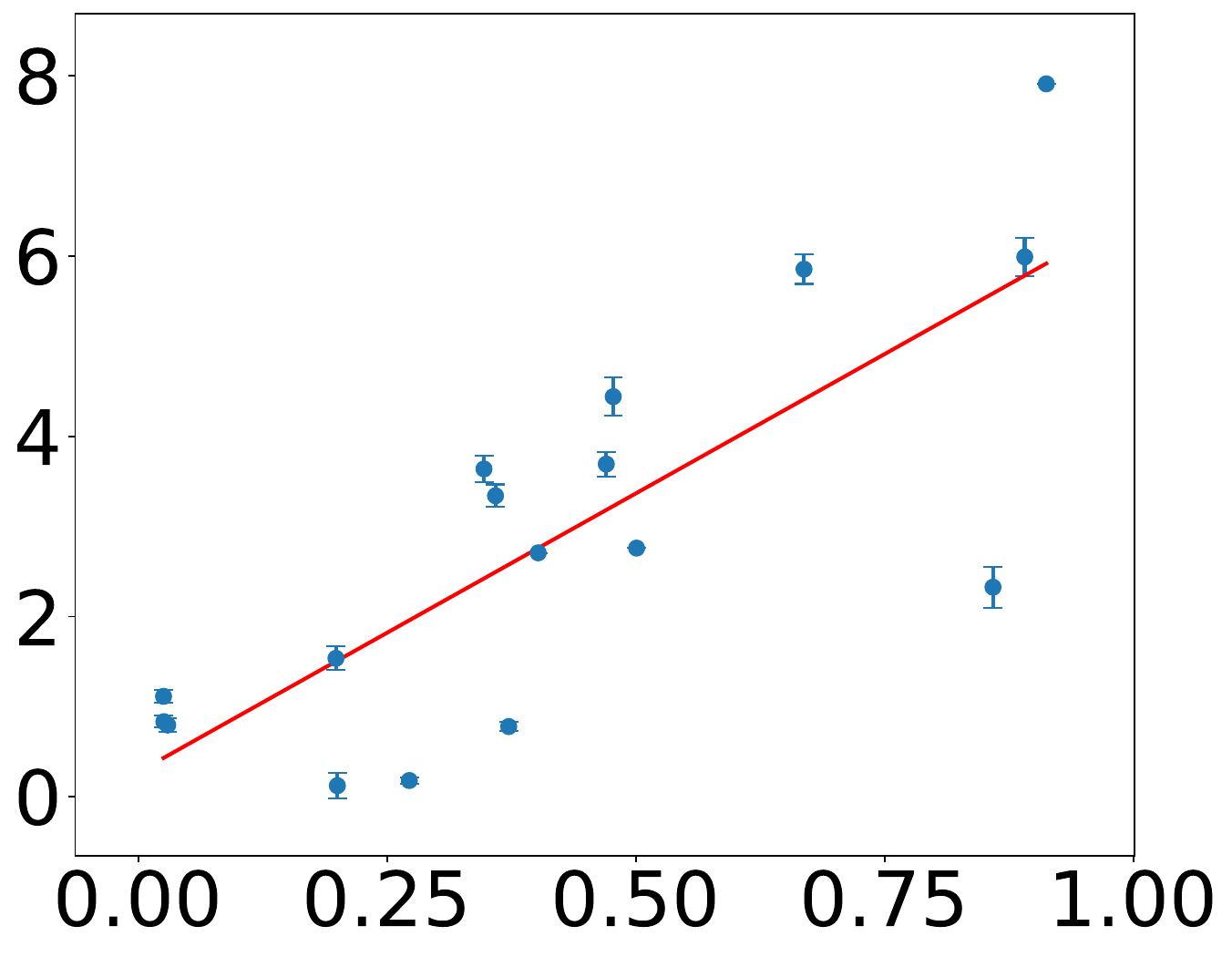}
        \caption{Llama-3-8B}\label{fig:kappa_llama}
    \end{subfigure}
    \begin{subfigure}{0.19\linewidth}
        \centering
        \includegraphics[width=\linewidth]{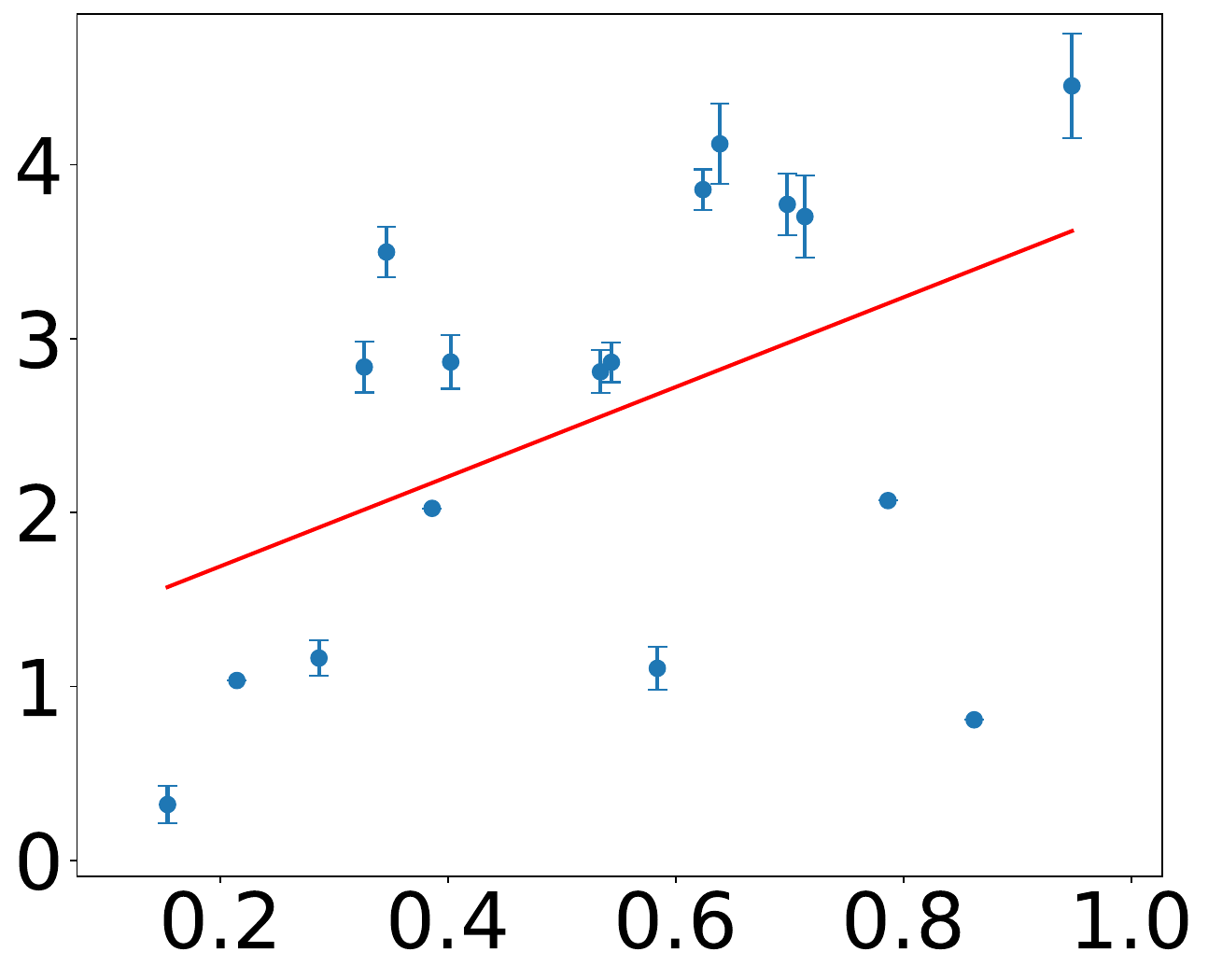}
        \caption{Llama-3-8B-Instruct}\label{fig:kappa_llama_inst}
    \end{subfigure}
    \begin{subfigure}{0.19\linewidth}
        \centering
        \includegraphics[width=\linewidth]{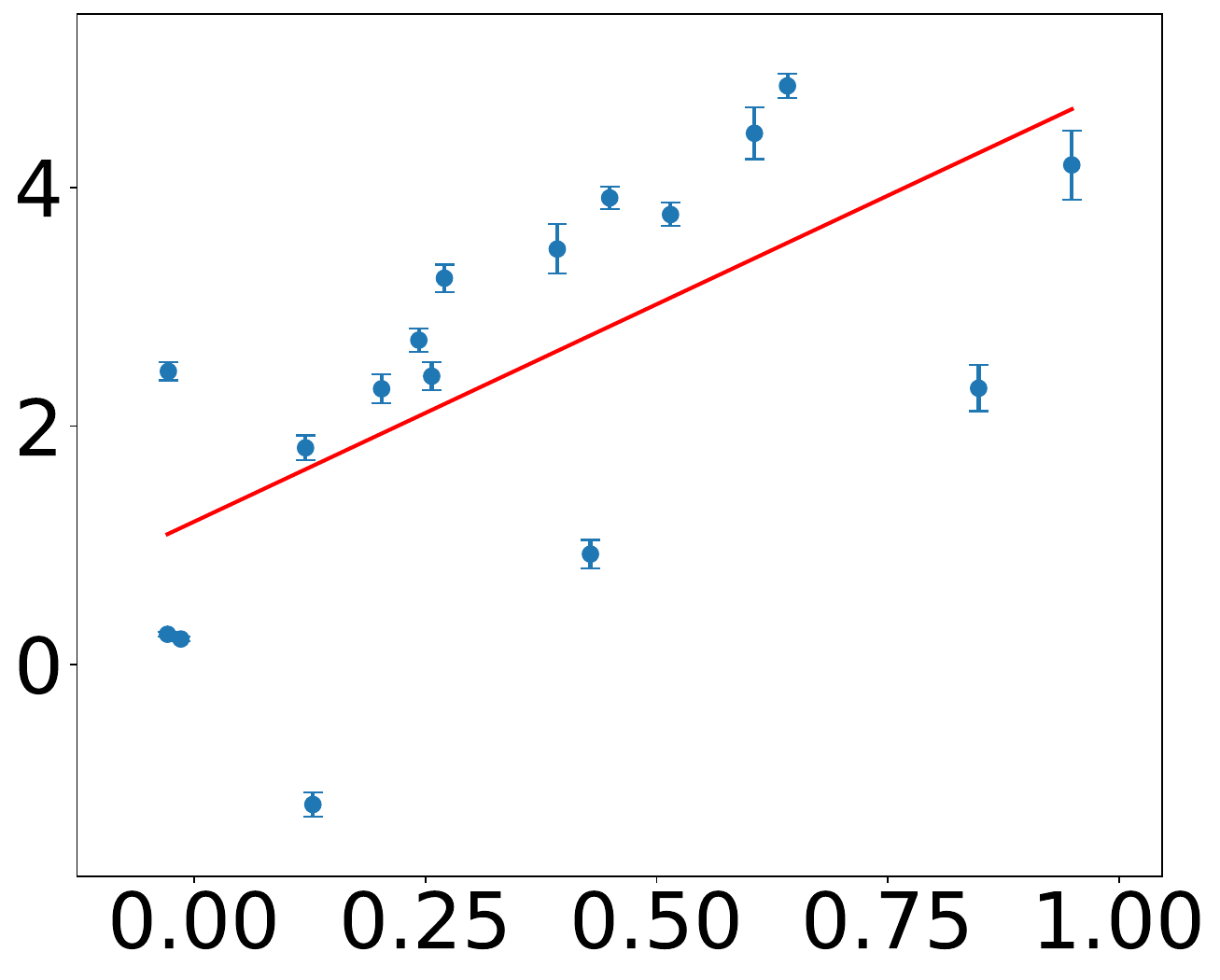}
        \caption{Mistral-7B-Instruct}\label{fig:kappa_mistral}
    \end{subfigure}

    \caption{Kappa vs. prediction depth difference. Each point in the plots represents a dataset; the y-axis denotes the average prediction depth difference, or difference in layers at which the model committed to answers, between correctly and incorrectly answered questions and the x-axis denotes Cohen's Kappa score, or accuracy adjusted for random chance. The positive correlations across the models show that as task competence increases, represented by the Kappa score, the degree to which uncertainty affects the inference dynamics increases, represented by the prediction depth difference. (\texttt{Llama-3-8B} and \texttt{Mistral-7B-Instruct} have $p < 0.05$ while \texttt{Llama-3-8B-Instruct} has $p = 0.06$. Other larger models do not observe significant trends.)}
    \label{fig:kappa}
\end{figure*}

%% file: src/table/pd_correlation.tex
\begin{table*}[htb!]  
\caption{Pearson correlation between answer incorrectness and PD. Greater values would mean that incorrectness correlates with later prediction commitment layers. Correlations are generally weak, with 80\% below 0.300, but with 97\% being positive. This consistency shows LLMs' potential to leverage uncertainty in their inference dynamics. (Values in parentheses show standard error rates. Values above 0.300 are bold.)}
\label{tab:pd-corr-table}
\vskip 0.15in
\centering
\resizebox{\textwidth}{!}{
\begin{sc}
\begin{tabular}{lccccc}
\toprule
Dataset & Llama-3-8B & Llama-3-8B-Instruct & Vicuna-13B & Mistral-7B-Instruct & Mistral-Nemo-Instruct \\
\midrule
ANLI-R1 & 0.192*(0.015) & 0.266*(0.013) & 0.195*(0.014) & 0.118*(0.014) & \textbf{0.428}*(0.012) \\
ANLI-R2 & 0.174*(0.016) & \textbf{0.321}*(0.013) & 0.194*(0.014) & 0.146*(0.014) & \textbf{0.464}*(0.011) \\
ANLI-R3 & 0.238*(0.015) & 0.258*(0.013) & 0.215*(0.013) & 0.284*(0.009) & 0.240*(0.013) \\
ARC-Easy & \textbf{0.449}*(0.011) & \textbf{0.351}*(0.019) & 0.069*(0.021) & \textbf{0.443}*(0.017) & \textbf{0.332}*(0.019) \\
ARC-Challenge & \textbf{0.373}*(0.017) & \textbf{0.363}*(0.017) & 0.040(0.030) & \textbf{0.322}*(0.018) & \textbf{0.357}*(0.017) \\
BoolQ & 0.061*(0.012) & 0.159*(0.014) & 0.058*(0.014) & -0.168*(0.014) & -0.083*(0.014) \\
BoolQ w/ context & 0.175*(0.011) & 0.132*(0.014) & 0.233*(0.013) & 0.107*(0.014) & -0.081*(0.014) \\
CommonsenseQA & \textbf{0.316}*(0.011) & 0.255*(0.009) & -0.045*(0.010) & 0.274*(0.009) & 0.258*(0.010) \\
HellaSwag & 0.010(0.012) & 0.047*(0.016) & 0.188*(0.014) & 0.261*(0.013) & \textbf{0.309}*(0.013) \\
LogiQA & 0.174*(0.014) & 0.120*(0.011) & 0.113*(0.014) & 0.216*(0.011) & 0.269*(0.015) \\
MMLU & \textbf{0.332}*(0.013) & 0.217*(0.008) & 0.139*(0.014) & 0.265*(0.009) & \textbf{0.364}*(0.013) \\
QASC & 0.287*(0.010) & 0.252*(0.010) & 0.129*(0.011) & 0.213*(0.011) & 0.248*(0.010) \\
QASC w/ context & \textbf{0.374}*(0.010) & 0.263*(0.010) & \textbf{0.408}*(0.009) & 0.298*(0.010) & 0.164*(0.011) \\
QuAIL & \textbf{0.318}*(0.013) & \textbf{0.337}*(0.009) & 0.276*(0.009) & \textbf{0.381}*(0.009) & \textbf{0.334}*(0.013) \\
RACE & \textbf{0.315}*(0.013) & \textbf{0.327}*(0.013) & 0.247*(0.013) & \textbf{0.370}*(0.009) & \textbf{0.396}*(0.012) \\
SciQ & \textbf{0.480}*(0.007) & 0.213*(0.009) & 0.123*(0.014) & \textbf{0.484}*(0.008) & 0.101*(0.014) \\
SciQ w/ context & 0.217*(0.014) & 0.083*(0.009) & 0.192*(0.014) & 0.286*(0.013) & 0.065*(0.014) \\
\bottomrule
\multicolumn{6}{r}{\scriptsize{$^{*}p < 0.05$}} \\
\end{tabular}
\end{sc}
}
\end{table*}

%% file: texes/2_related_works.tex
\subsection{Logit Lens and Tuned Lens}
To analyze the inference dynamics of models, we leverage Tuned Lens~\cite{belrose2023eliciting}, a variant of the method of the Logit Lens~\cite{nostalgebraist2020logit}. The original Logit Lens is an early-stage technique that decodes intermediate activations into vocabulary space using the classification head. It has been leveraged to gain insights into how LLMs process and generate languages~\cite{wang2025extending}, including analysis on reasoning pathways~\cite{li2024understanding}, multilingual representations~\cite{wendler2024llamas}, and safety alignment~\cite{golgoon2024mechanistic}. Building upon this, \citet{belrose2023eliciting} proposed Tuned Lens, a refinement of Logit Lens, which trains an affine probe for each block in a frozen pre-trained model. This method offers more reliable interpretations of model predictions across layers.

\subsection{Prediction Depth}
Prediction depth is the earliest layer in a deep learning model where the highest-probability prediction matches the final output layer's prediction, differs from the previous layer's prediction, and remains the same through subsequent layers. It can be interpreted as the commitment layer or the effective number of layers used for inference. \citet{baldock2021deep} first defined and validated it as a measure of computational difficulty for individual examples in computer vision tasks. However, there is limited empirical research on PD in LLMs. While \citet{belrose2023eliciting} has tested how effectively PD correlates with iteration learned on a single model, we analyze it with relation to epistemic uncertainty across a variety of models. 

\subsection{Inference Dynamics}
The term ``\emph{inference dynamics}'' as used in this work is an umbrella term that refers to the patterns that arise at inference, namely prediction probability trajectories and prediction refinement across layers. Studies in this field have found conflicting results for how a model produces its final output, with some results showing a gradual refinement across layers while others have found layers disproportionately contributing to the output and inducing sudden ``\emph{decisions}'' \cite{nostalgebraist2020logit, kongmanee2025attempt}. While the research is inconclusive, methods for early-exiting still try to leverage the knowledge we have on inference dynamics to reduce computational cost by using predictions at intermediate layers \cite{xin-etal-2020-deebert, schuster2022confident, varshney-etal-2024-investigating}. Our results related to PD and prediction trajectories would help advance these endeavors. 

%% file: texes/5_conclusion.tex
In this paper, we analyze if epistemic uncertainty causes large language models (LLMs) to exhibit different inference dynamics.
Using a Logit Lens-based method, we examine layer-wise token probability trajectories and prediction depth (PD) statistics of 5 LLMs across 11 datasets.
We demonstrate that LLMs exhibit similar token probability dynamics and tend to commit to final decisions at specific layers largely independent of output uncertainty.
This implies that inference is generally characterized by abrupt decision-making unaffected by levels of uncertainty.
In addition, we observe that models exhibit the potential for greater adaptability in their inference dynamics as PD shows weak but consistent correlations with epistemic uncertainty. 
Further, models exhibit greater degrees of inference adaptability with respect to uncertainty at tasks they are more proficient at, implying greater degrees of adaptability could emerge in more competent models.
These findings have implications for interpretability methods assessing uncertainty at inference as well as for adaptive inference techniques (\ie test-time scaling, early exiting) that leverage intermediate prediction confidence. Further testing should be done on more models to validate the results of this exploratory work, especially on more competent models. More rigorous tests that include analyses using varying levels of uncertainty and accounts for hallucination on known knowledge is needed as well. Additionally, similar experiments for aleatoric uncertainty may be done for a more thorough investigation. 

%% file: texes/appendix.tex
\section{Broader Impacts Statement}
\input{texes/impact_statement}

\section{Reproducibility Statement}
\subsection{Prompts}
Question prompts from the original MCQ datasets were augmented with directions to answer with a single letter. If supplementary context that is not necessarily needed to answer the question is provided, a separate version of the questions with contexts were also tested. Note the use of brackets around the label in the directions and after the `\emph{Answer: }' to prompt the model to output a letter label. The format is as following:

\begin{tcolorbox}[breakable, enhanced, top=1pt, left=1pt, right=1pt, bottom=1pt]
    Answer the question with a single letter like [A]. 
    
    Mesophiles grow best in moderate temperature, typically between 25°C and 40°C (77°F and 104°F). Mesophiles are often found living in or on the bodies of humans or other animals. The optimal growth temperature of many pathogenic mesophiles is 37°C (98°F), the normal human body temperature. Mesophilic organisms have important uses in food preparation, including cheese, yogurt, beer and wine.
    
    What type of organism is commonly used in preparation of foods such as cheese and yogurt?
    
    A. viruses
    
    B. protozoa
    
    C. gymnosperms
    
    D. mesophilic organisms
    
    Answer: [
\end{tcolorbox}

\subsection{Sample Size}
The model did not produce relevant answer labels for every question. For example, it sometimes tried to answer with a sentence such as ``\emph{The answer is $\cdots$.}'' Hence, we consider only the questions that were answered with a single label token. The filtering results in sample sizes that are different across model and dataset pairs as is shown in Table~\ref{tab:sensical}.

\input{src/table/sensical}

\clearpage

\subsection{Tuned Lens Training}

\begin{wraptable}{R}{0.25\textwidth}
\vspace{-0.3in}
\caption{Training details for Tuned Lens}
\label{tab:hyperparameters}
\centering
\small
\begin{sc}
\begin{tabular}{lr}\\ \toprule  
Hyperparameters & Values \\\midrule
Optimzier & Adam \\
Learning rate & 0.001 \\
Training steps & 1000 \\
Weight decay & 0.01 \\
Momentum & 0.9 \\
Tokens per step & $2^{18}$ \\ \bottomrule
\end{tabular}
\end{sc}
\end{wraptable}

We trained Tuned Lens for the \texttt{Mistral-7B-Instruct-v0.1} and \texttt{Mistral-Nemo-Instruct-2407} using WikiText-103-dataset~\cite{merity2017pointer}, following the original implementations on GitHub~\footnote{\url{https://github.com/AlignmentResearch/tuned-lens}}. 
We use Intel(R) Xeon(R) Silver 4214R CPU @ 2.40GHz (24 cores), 188GiB System memory, and four 48GB NVIDIA Quadro RTX 8000 GPUs.
Table~\ref{tab:hyperparameters} shows datasets and hyperparameters for training Tuned Lenses.

%% file: texes/impact_statement.tex
This paper presents work whose goal is to advance the field of machine learning.
There are many potential societal consequences of our work, none of which we feel must be specifically highlighted here.

%% file: src/table/sensical.tex
\begin{table*}[htb!] 
\caption{Number of sensical answers for model and dataset pairs.}
\label{tab:sensical}
\vskip 0.15in
\centering
\resizebox{\textwidth}{!}{
\begin{sc}
\begin{tabular}{lccccc}
\toprule
Dataset & Llama-3-8B & Llama-3-8B-Instruct & Vicuna-13B & Mistral-7B-Instruct & Mistral-Nemo-Instruct \\
\midrule
ANLI-R1 & 3909 & 5000 & 5000 & 5000 & 4996 \\
ANLI-R2 & 3787 & 5000 & 4998 & 5000 & 5000 \\
ANLI-R3 & 4037 & 5000 & 5000 & 10000 & 5000 \\
ARC-Easy & 4935 & 2241 & 2206 & 2151 & 2237 \\
ARC-Challenge & 2556 & 2567 & 1116 & 2557 & 2531 \\
BoolQ & 7479 & 5000 & 5000 & 4999 & 5000 \\
BoolQ w/ context & 7481 & 5000 & 5000 & 5000 & 5000 \\
CommonsenseQA & 6372 & 9737 & 9741 & 9706 & 9325 \\
HellaSwag & 7394 & 3772 & 4988 & 5000 & 4720 \\
LogiQA & 4687 & 7375 & 4996 & 7373 & 3745 \\
MMLU & 4969 & 14032 & 4991 & 9822 & 4742 \\
QASC & 7952 & 8122 & 8134 & 8112 & 8062 \\
QASC w/ context & 7917 & 8134 & 8134 & 8132 & 7998 \\
QuAIL & 4996 & 9999 & 10246 & 10000 & 5000 \\
RACE & 4981 & 4998 & 5000 & 9999 & 4975 \\
SciQ & 11679 & 11679 & 5000 & 10000 & 4996 \\
SciQ w/ context & 4972 & 11679 & 5000 & 4999 & 4992 \\
\bottomrule
\end{tabular}
\end{sc}
}
\end{table*}